\documentclass[10pt,twocolumn,letterpaper]{article}
\usepackage{cvpr}
\usepackage{times}
\usepackage{epsfig}
\usepackage{graphicx}
\usepackage{amsmath}
\usepackage{amssymb}
\usepackage{hyperref}
\usepackage{fixltx2e}

\usepackage{cite}

\def\cvprPaperID{} 
\def\confName{CVPR}
\def\confYear{2023}

\begin{document}

\title{Image-to-LaTeX Converter for Mathematical Formulas and Text}

\author{
    Daniil Gurgurov\textsuperscript{1} \quad Aleksey Morshnev\textsuperscript{2} \\
    \small \textsuperscript{1}Department of Language Science and Technology, Saarland University \\
    \small \textsuperscript{2}Department of Data Science and Artificial Intelligence, Saarland University \\
    \small \texttt{\{dagu00001, almo00008\}@stud.uni-saarland.de}
}

\maketitle

\begin{abstract}
In this project, we train a vision encoder-decoder model to generate LaTeX code from images of mathematical formulas and text. Utilizing a diverse collection of image-to-LaTeX data, we build two models: a base model with a Swin Transformer encoder and a GPT-2 decoder, trained on machine-generated images, and a fine-tuned version enhanced with Low-Rank Adaptation (LoRA) trained on handwritten formulas. We then compare the BLEU performance of our specialized model on a handwritten test set with other similar models, such as \href{https://github.com/lukas-blecher/LaTeX-OCR}{Pix2Text}, \href{https://github.com/OleehyO/TexTeller/tree/main}{TexTeller}, and \href{https://github.com/hoang-quoc-trung/sumen}{Sumen}. Through this project, we contribute open-source models for converting images to LaTeX and provide from-scratch code for building these models with distributed training and GPU optimizations.
\end{abstract}

\section{Introduction}

Transforming images containing text and formulas into a readable format presents significant challenges in the field of document digitalization \cite{yuan2017multi, nishizawa2020mathseer, peng2021mathbert, peng2021image, 9672359, zhou2022end, chen2023printed}. Traditional Optical Character Recognition (OCR) systems have relied on Convolutional Neural Networks (CNNs) \cite{fukushima1980neocognitron} for image processing and Recurrent Neural Networks (RNNs) \cite{rumelhart1986learning} for text generation \cite{mishra2020extraction, memon2020handwritten, parthiban2020optical}. However, recent advancements in Transformer-based models have demonstrated significant potential in improving the performance of OCR tasks.

This project explores the implementation of a vision encoder-decoder model based on the Swin Transformer \cite{liu2021swin} and GPT-2 \cite{radford2019language}, following the architecture proposed in TrOCR \cite{li2023trocr}. The TrOCR model, which stands for Transformer-based Optical Character Recognition, utilizes pre-trained Transformer models for both the encoder and decoder components \cite{vaswani2017attention}. The encoder processes the image by dividing it into patches and applying a Transformer network, while the decoder generates the text output using a pre-trained text Transformer model. 

By adopting this architecture, we aim to achieve effective performance in converting formulas from both high-quality computer-generated and handwritten images into LaTeX text. We will compare our results on handwritten formulas with other models of this type, such as \href{https://github.com/lukas-blecher/LaTeX-OCR}{Pix2Text} \cite{pix2text}, \href{https://github.com/OleehyO/TexTeller/tree/main}{TexTeller} \cite{texteller}, and \href{https://github.com/hoang-quoc-trung/sumen}{Sumen} \cite{sumen}, to evaluate the relative performance and robustness of our approach.

Our code and pre-trained models are publicly available on GitHub\footnote{\url{https://github.com/d-gurgurov/im2latex}} and HuggingFace\footnote{\url{https://huggingface.co/DGurgurov/im2latex}}. This open-source release aims to support further research and development in the field of OCR, particularly for mathematical and scientific documents, by providing a from-scratch implementation of distributed training and GPU optimization techniques and a 240M base model for converting images to LaTeX code.

\section{Related Work}

\subsection{Traditional OCR Methods}

The field of OCR has seen significant progress with the introduction of deep learning techniques \cite{subramani2020survey}. Traditional methods have relied on CNNs for feature extraction from images and RNNs for sequential text generation \cite{mishra2020extraction, memon2020handwritten, parthiban2020optical}. While effective, these approaches often require additional language models to improve accuracy and involve sophisticated pre/post-processing steps.

\subsection{Transformer-Based OCR Models}

The TrOCR model advances traditional methods by using a Transformer-based architecture for both image and text processing. It employs a Vision Transformer (ViT) \cite{touvron2021training} in the encoder to extract visual features and a textual Transformer, such as RoBERTa \cite{liu2019roberta}, in the decoder for generating text. This end-to-end approach leverages pre-trained models for superior text recognition performance.

Ablation studies explored various encoder-decoder combinations, including DeiT \cite{touvron2021trainingdataefficientimagetransformers}, BEiT \cite{beit}, and ResNet-50 \cite{he2015deepresiduallearningimage} for encoders, and RoBERTa$_{BASE}$ and RoBERTa$_{LARGE}$ for decoders. BEiT encoders with RoBERTa$_{LARGE}$ decoders achieved the best performance, surpassing CRNN \cite{DBLP:journals/corr/ShiBY15} and Tesseract \cite{4376991} models, with BEiT$_{BASE}$ and RoBERTa$_{LARGE}$ reaching an F1 score of 79.36\% on the SROIE dataset \cite{DBLP:journals/corr/abs-2103-10213}. 

Additional ablation experiments verified the positive impact of pre-trained model initialization, data augmentation, and two stages of pre-training on the TrOCR models. Starting from a scratch model, incremental improvements were observed with each enhancement: using a pre-trained model, applying data augmentation, and conducting two-stage pre-training, ultimately achieving an F1 score of 95.84\%.

\subsection{Swin Transformer and GPT-2}

The Swin Transformer, introduced by \cite{liu2021swin}, extends the capabilities of the ViT by incorporating hierarchical feature maps and shifted windows, enhancing its ability to capture local and global image context. This makes it particularly suitable for processing complex document images, such as those containing LaTeX formulas.

GPT-2 \cite{radford2019language}, a powerful open-source language model, is known for its ability to generate coherent and contextually relevant text. By integrating GPT-2 as the decoder in our system, we aim to accurately transcribe the extracted information into LaTeX code.

\subsection{Existing Models}

Several models have been developed to address the task of converting mathematical formulas to LaTeX code, which are Pix2Text, TexTeller, and Sumen-base. \textit{Pix2Tex} employs a vision encoder-decoder architecture similar to TrOCR. It uses a ResNet backbone as the encoder and a Transformer decoder. The model is trained on a dataset of approximately 100,000 image-formula pairs, primarily consisting of computer-generated images. Pix2Tex has around 25 million parameters. \textit{TexTeller} also adopts a TrOCR-like architecture but utilizes a Vision Transformer (ViT) as the encoder and a Transformer decoder. It is reportedly trained on a larger dataset of over 7.5 million image-formula pairs, which includes both computer-generated and a small portion of handwritten formulas. The exact number of parameters is not publicly disclosed, but it has 300 million parameters. \textit{Sumen} takes a different approach by using a Swin Transformer as the encoder and a GPT-2 model as the decoder. This architecture is similar to our proposed model. Sumen is trained on a dataset of approximately 6.9 million image-formula pairs, with a mix of computer-generated and handwritten samples. The model has about 350 million parameters. 

It's important to note that while these models have shown promising results, the exact details of their training data and architectures are not always fully disclosed or verified. Our work aims to provide a more transparent and comprehensive approach, with clear documentation of the model architecture, training data composition, and experimental results.

\section{Methodology}

Our approach to developing an Image-to-LaTeX converter involves a two-step training process, similar to the one introduced by TrOCR. First, we train a base model on images of printed formulas. This model is subsequently fine-tuned on a dataset containing handwritten formulas. The following sections provide a detailed description of each step and the necessary sub-steps.

\subsection{Step 1: Training the Base Model on Printed Formulas}

\subsubsection{Data Preparation}
The initial step involves preparing a dataset of printed formulas. We utilize a publicly available pre-processed dataset that provides pairs of images containing printed mathematical formulas and their corresponding LaTeX code\footnote{\url{https://huggingface.co/datasets/linxy/LaTeX_OCR}}. This dataset is split into training, validation, and test sets in an 80:10:10 ratio, with a fixed seed for reproducibility and further comparisons with other models, resulting in the following numbers: train - 441,872, validation - 55,234, and test - 55,234.

The dataset preparation process involved cleaning approximately 1 million LaTeX formula image-text pairs initially scraped from arXiv. This raw dataset, termed "raw\_formulas," underwent extensive cleaning steps to remove irrelevant data and overly complex formulas. Specifically, formulas with aspect ratios greater than 0.8 and character lengths exceeding 200 were removed. Additionally, specific LaTeX commands and environments such as \texttt{\textbackslash tag}, \texttt{\textbackslash text}, and equation environments were removed or standardized. The cleaned dataset, integrated with the im2latex-100K dataset\footnote{\url{https://huggingface.co/datasets/yuntian-deng/im2latex-100k-raw}}, resulted in a combined dataset of 550K formula-image pairs, termed "cleaned\_formulas," and was used for training our image-to-LaTeX converter model.

\subsubsection{Model Architecture}
\begin{figure}[h]
\centering
\includegraphics[width=0.5\textwidth]{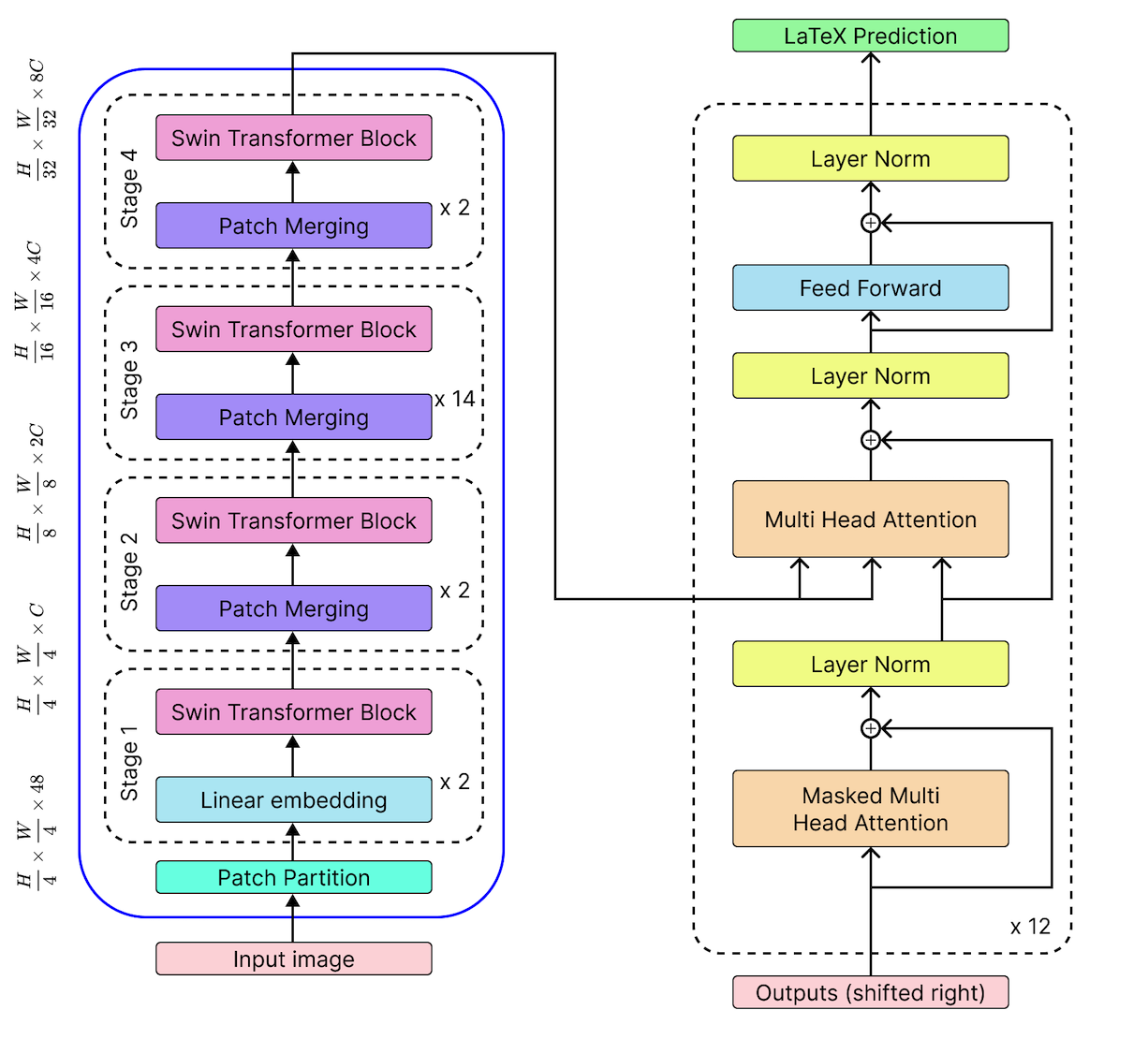}
\caption{Proposed architecture for training a base formula recognition model.}
\label{fig:base_model}
\end{figure}

We employ a Vision Encoder-Decoder Model for our task, combining a Swin Transformer (\textit{microsoft/swin-base-patch4-window7-224-in22k}) as the encoder and GPT-2 (\textit{gpt2-base}) as the decoder, resulting in 243,433,656 total parameters. The encoder processes the input image to extract visual features, while the decoder generates the corresponding LaTeX code. We use the tokenizer and feature extractor provided by the Hugging Face library \cite{wolf-etal-2020-transformers}. The tokenizer converts LaTeX code into a sequence of tokens, while the feature extractor processes images into a format suitable for the encoder. The architecture of the base model is illustrated in Figure \ref{fig:base_model}.

\subsection{Step 2: Fine-Tuning on Handwritten Formulas}

\subsubsection{Data Preparation}

For fine-tuning the model, we utilize a dataset containing images of handwritten formulas paired with their corresponding LaTeX representations. This approach ensures our model can effectively handle both printed and handwritten text inputs. The handwritten formulas dataset is sourced from the LaTeX\_OCR project\footnote{\url{https://huggingface.co/datasets/linxy/LaTeX_OCR}}, specifically the subset of \textit{Human Handwritten Formulas}: 1,338 examples (train - 1,200, validation - 68, test - 70) of formulas handwritten on electronic screens, primarily sourced from CROHME \cite{mouchere2016icfhr2016}.

\subsubsection{Model Architecture}

Our architecture for fine-tuning integrates an adapter approach, specifically LoRA (Low-Rank Adaptation of Large Language Models) \cite{hu2021loralowrankadaptationlarge}, into the Vision Encoder-Decoder Model trained in the first step. Utilizing this parameter-efficient fine-tuning method, the total number of trainable parameters is reduced to 3,096,576.

LoRA introduces adaptability within specific modules critical to the adaptation process:

\begin{itemize}
\item For the encoder (Swin Transformer), LoRA targets the attention modules (\texttt{attn.qkv}, \texttt{attn.proj}) and the MLP layers (\texttt{mlp.fc1}, \texttt{mlp.fc2}).
\item In the decoder (GPT-2), the adaptation focuses on the cross-attention module (\texttt{c\_attn}), projection layers (\texttt{c\_proj}), fully connected layers (\texttt{c\_fc}), and the attention module (\texttt{attn.c\_proj}).
\end{itemize}

These targeted modules are configured with LoRA parameters, including a reduction factor (\texttt{r}) of 16, scaling factor (\texttt{lora\_alpha}) of 8, and dropout rate (\texttt{lora\_dropout}) of 0.2. The adapter modules are integrated seamlessly into the base model, enhancing its capability to learn from the handwritten formula dataset without extensive modification of the original architecture.

\section{Experiments}

\subsection{Training Configuration}

For both Step 1 (Training the Base Model on Printed Formulas) and Step 2 (Fine-Tuning on Handwritten Formulas), the training configurations are largely identical. The model is trained using PyTorch's Distributed Data Parallel (DDP) \cite{li2020pytorchdistributedexperiencesaccelerating} for efficient multi-GPU processing. We utilize the AdamW \cite{loshchilov2019decoupledweightdecayregularization} optimizer with a linear learning rate scheduler that includes warmup steps to stabilize training.

The datasets are managed using custom classes (\texttt{LatexDataset} and specific collators) to handle image preprocessing, LaTeX sequence tokenization, batching, and padding. Key hyperparameters for the first and second steps respectively include:

\begin{itemize}
\item Batch Size: 32 for both training and validation.
\item Learning Rate: \{1e-4, 2e-4\}, adjusted dynamically using a scheduler.
\item Gradient Clipping: Applied to prevent gradient explosions with a maximum norm of 1.0.
\item Epochs: \{10, 40\}, iterating over the dataset for multiple passes.
\item Evaluation Steps: The model is evaluated periodically every \{200, 40\} steps on the validation set using BLEU scores to monitor performance.
\end{itemize}

\subsection{Training Loop}

The training loop for both steps involves iterating over the dataset for a specified number of epochs. During each epoch, the model performs a forward pass on the batched data, computes the loss, and applies backpropagation to update the model weights. Moreover, gradient accumulation is implemented to combine gradients over multiple mini-batches before performing a weight update. This technique effectively increases the batch size without requiring additional GPU memory, which is particularly useful for training large models.

Gradient clipping ensures stable training, preventing gradients from exceeding a specified norm. Checkpoints are saved based on the best validation loss observed during training to retain the optimal model configuration. After training, the best-performing model is evaluated on the respective test set to compute final test loss and BLEU scores, assessing the model's generalization ability.

\subsection{GPU Optimization}

To maximize computational efficiency and accelerate training, we employed several GPU optimization techniques using 4 NVIDIA H100 GPUs. This setup allowed us to process 4 batches of 32 examples simultaneously, effectively handling 128 examples per iteration. The further following optimizations were implemented:

\begin{itemize}
\item \textbf{Reduced Float Precision:} We adjusted the default precision from "highest" to "high" using PyTorch's float32 matrix multiplication precision setting. This setting affects how float32 matrix multiplications are computed. "highest" uses the full float32 datatype (24 mantissa bits with 23 bits explicitly stored) for internal computations. "high" employs either the TensorFloat32 datatype (10 mantissa bits explicitly stored) or treats each float32 number as the sum of two bfloat16 numbers (approximately 16 mantissa bits with 14 bits explicitly stored).
  
\item \textbf{Model Compilation with Torch:} We utilized PyTorch's model compilation feature to optimize our model. This compilation process transforms the PyTorch model into optimized kernels, which can significantly improve execution speed. The compiled model is functionally equivalent to the original but benefits from various backend optimizations, including kernel fusion and memory layout optimizations.

\item \textbf{Automatic Mixed Precision (AMP):} We implemented Automatic Mixed Precision training using PyTorch's \texttt{autocast} context manager. AMP automatically chooses the appropriate precision for each operation, using float16 or bfloat16 where possible to speed up computations, while maintaining float32 precision where necessary for numerical stability. This approach offers a balance between computational efficiency and training stability.
\end{itemize}

These GPU optimization techniques collectively allowed us to train our large-scale model more efficiently, reducing memory usage and training time while maintaining model performance. The use of H100 GPUs, known for their high memory bandwidth and computational power, further enhanced the effectiveness of these optimizations.

\subsection{Results}

In this section, we present the performance comparison of our model, Im2Latex, with other state-of-the-art models: TexTeller, Pix2Tex, and Sumen. The evaluation metric used is the Google BLEU score \cite{wu2016googles}, which measures the quality of LaTeX code generation from images. Our results show that Im2Latex (ours) achieves a BLEU score of 0.67, outperforming Pix2Tex and Sumen but falling behind TexTeller. Table \ref{tab:results} summarizes the performance of each model along with their respective training dataset sizes and parameter counts.

\begin{table}[h]
\centering
\begin{tabular}{|l|c|c|c|}
\hline
\textbf{Model} & \textbf{Google BLEU} & \textbf{Data Size} & \textbf{Parameters} \\
\hline
Im2Latex & \textit{0.67} & 441K & 243M \\
TexTeller & \textbf{0.77} & 7.5M & 300M \\
Pix2Text & 0.07 & 100K & 25M \\
Sumen & 0.47 & 6.9M & 350M \\
\hline
\end{tabular}
\caption{Comparison of performance and dataset sizes for various image-to-LaTeX models.}
\label{tab:results}
\end{table}

Training and validation curves for Im2Latex are provided in Figures \ref{fig:base} and \ref{fig:lora} of the appendix for a more detailed analysis of the model's performance over training epochs.

\section{Conclusion}

In this study, we introduced Im2Latex, a vision encoder-decoder model designed for converting images of mathematical formulas into LaTeX code. Our model achieves a Google BLEU score of 0.67, demonstrating competitive performance relative to other models such as TexTeller, Pix2Tex, and Sumen. Despite this, Im2Latex's performance is comparable to larger models like TexTeller and Sumen, which have significantly larger training datasets and parameter counts.

However, it is important to note that the fairness of these comparisons may be influenced by the potential overlap of training and test images in the datasets used by other models. The possibility of dataset contamination could impact the validity of the performance evaluations and should be considered when interpreting the results.

Our work contributes to the field by providing both the model and its training code as open-source resources. This includes detailed implementations of distributed training and GPU optimization techniques, as well as a 240M base model for image-to-LaTeX conversion. By making these resources publicly available, we aim to support further research and development in OCR and mathematical document analysis.

\bibliographystyle{ieee_fullname}
\bibliography{custom}

\newpage
\section*{Appendix}

\begin{figure}[h]
\centering
\includegraphics[width=0.5\textwidth]{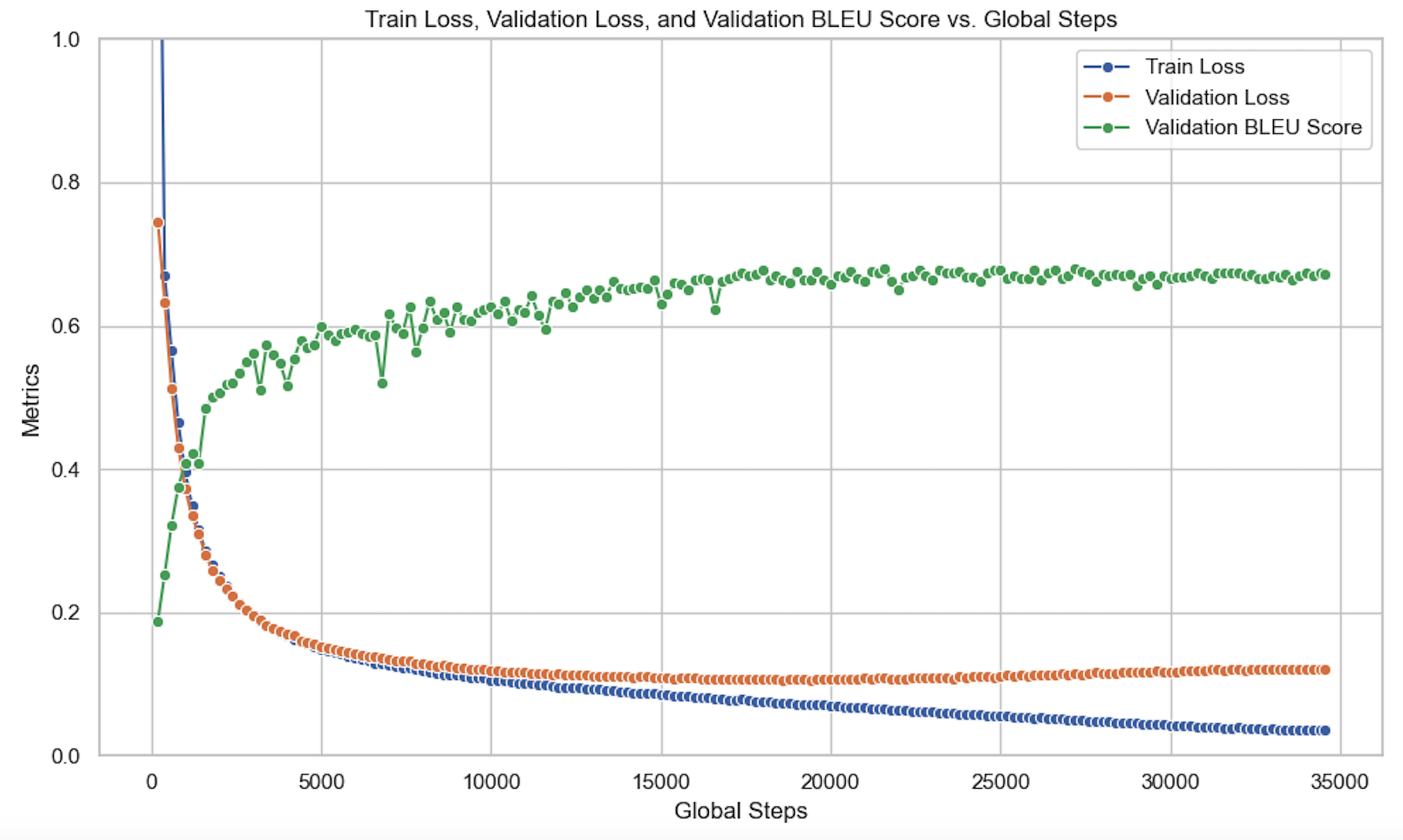}
\caption{Base-Model Training Details.}
\label{fig:base}
\end{figure}

\begin{figure}[h]
\centering
\includegraphics[width=0.5\textwidth]{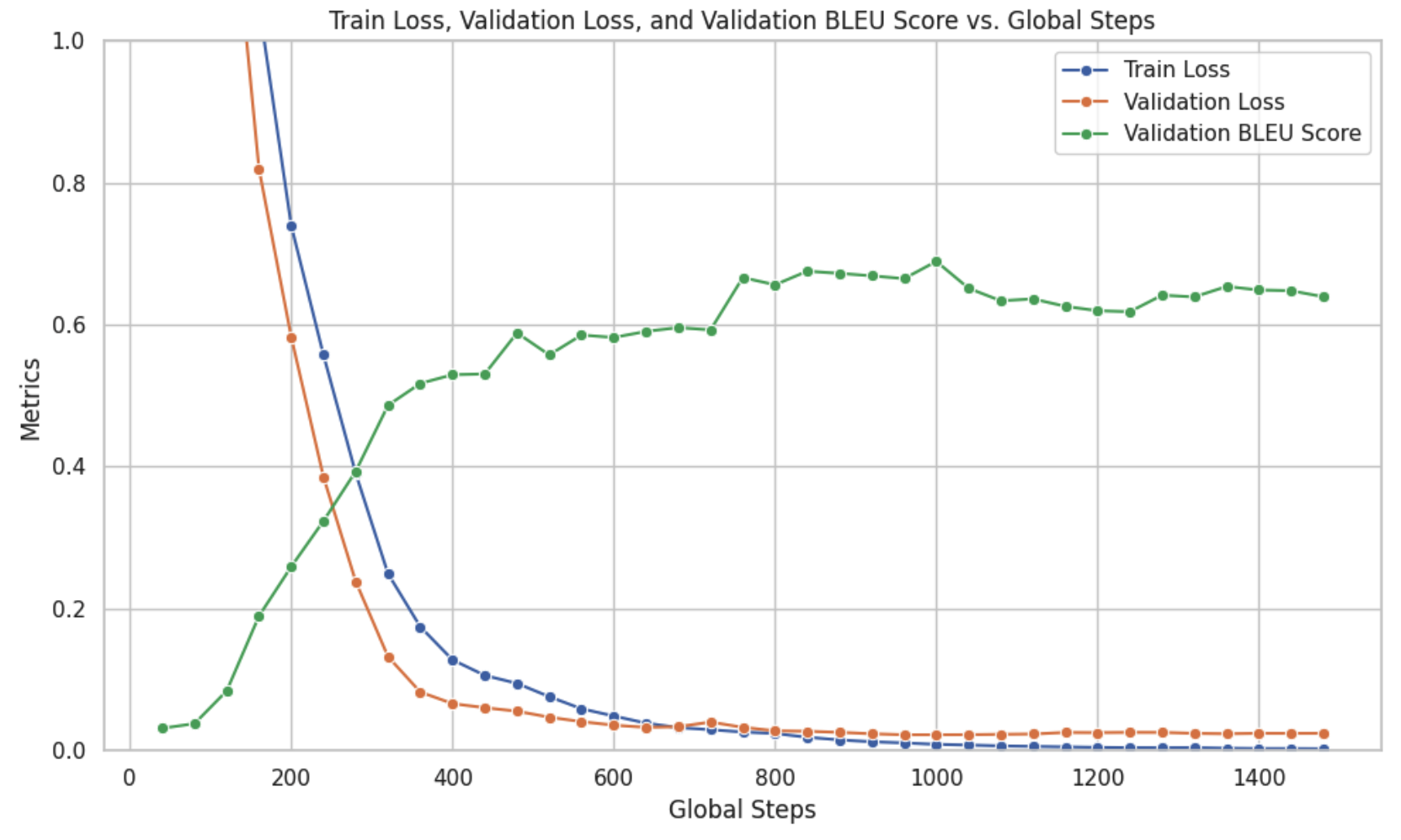}
\caption{LoRa-Model Training Details.}
\label{fig:lora}
\end{figure}

\end{document}